# DreamerV3 for Traffic Signal Control: Hyperparameter Tuning and Performance

Qiang Li [a], Yinhan Lin [a], Qin Luo [a] and Lina Yu [a, 1]

[a] *College of Urban Transportation and Logistics, Shenzhen Technology University, Shenzhen, Guangdong 518118, China*

**Abstract.** Reinforcement learning (RL) has evolved into a widely investigated technology for the development of smart TSC strategies. However, current RL algorithms necessitate excessive interaction with the environment to learn effective policies, making them impractical for large-scale tasks. The DreamerV3 algorithm presents compelling properties for policy learning. It summarizes general dynamics knowledge about the environment and enables the prediction of future outcomes of potential actions from past experience, reducing the interaction with the environment through imagination training. In this paper, a corridor TSC model is trained using the DreamerV3 algorithm to explore the benefits of world models for TSC strategy learning. In RL environment design, to manage congestion levels effectively, both the state and reward functions are defined based on queue length, and the action is designed to manage queue length efficiently. Using the SUMO simulation platform, the two hyperparameters (training ratio and model size) of the DreamerV3 algorithm were tuned and analyzed across different OD matrix scenarios. We discovered that choosing a smaller model size and initially attempting several medium training ratios can significantly reduce the time spent on hyperparameter tuning. Additionally, we found that the approach is generally applicable as it can solve two TSC task scenarios with the same hyperparameters. Regarding the claimed data-efficiency of the DreamerV3 algorithm, due to the significant fluctuation of the episode reward curve in the early stages of training, it can only be confirmed that larger model sizes exhibit modest data-efficiency, and no evidence was found that increasing the training ratio accelerates convergence.

## 1. Introduction

Traffic Signal Control (TSC) systems play a crucial role in managing traffic flow at signalized intersections and hold significant potential for alleviating traffic congestion and enhancing travel efficiency on urban roads. Through continuous research efforts, Deep reinforcement learning (RL) has emerged as a widely explored technology for the development of intelligent TSC strategies. It has demonstrated considerable suitability for handling complex and dynamic traffic environments [1-4].

However, current RL algorithms necessitate excessive interaction with the environment to learn effective policies, rendering them impractical for large-scale tasks.

---
[1] Corresponding Author: Lina Yu, yulina@sztu.edu.cn.

Recently, modern world models have shown remarkable potential for data-efficient learning in simulated environments and video games [5-7]. DreamerV3, the latest version of these models, has been proven to be the first algorithm to collect diamonds in Minecraft from scratch without human data or curricula.

DreamerV3 possesses appealing characteristics for policy learning [7]. It compiles general dynamics knowledge about the environment and allows for the prediction of future outcomes of potential actions from past experience, which reduces the amount of interaction with the environment by enabling imagination training. Moreover, it is a general and scalable algorithm that outperforms previous approaches across a wide range of domains with fixed hyperparameters. Only two hyperparameters, the training ratio and model size, require tuning. An extensive evaluation reveals that higher training ratios lead to substantially improved data efficiency, and larger models achieve not only higher final performance but also better data efficiency. Finally, in the representation of the environmental state, predictive information (recurrent state $h_t$ in RNN model) is incorporated, which contains richer data than mere historical trends alone. This inclusion aids in accelerating the learning process and enhancing the efficacy of traffic congestion control.

Although world models hold great promise, learning accurate world models for the TSC problem remains a significant open challenge. In this paper, we train a corridor TSC model using the DreamerV3 algorithm, in order to explore the benefits of world models for TSC strategy learning. The key contributions of this paper are summarized as follows:

- A corridor traffic signal control (TSC) model was trained using the DreamerV3 algorithm, and the benefits of world models for TSC strategy learning were explored.

- The two hyperparameters of the DreamerV3 algorithm—training ratio and model size—were tuned and analyzed. It was discovered that using a model size of S and experimenting with several medium training ratios can decrease the time required for hyperparameter tuning. Furthermore, as the model size increases, the range of appropriate training ratios becomes narrower. Additionally, when the model size is relatively large, both excessively small and excessively large training ratios are more likely to result in overfitting.

- It was proved through experiments that the proposed TSC method is successful in managing and preventing congestion in urban corridors and can improve traffic flow even in situations with significant traffic demand.

## 2. Dreamer V3

DreamerV3 is composed of two parts: World Model Learning and Actor Critic Learning [6-8].

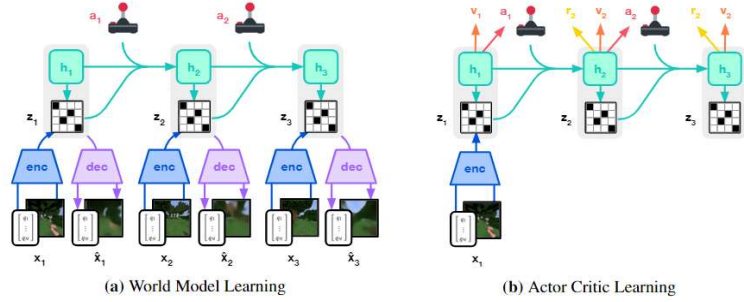

**Figure 1.** The DreamerV3 algorithm [7]

DreamerV3 learns a world model to obtain rich representations of the environment and enable imagination training by predicting future representations and rewards for potential actions. The world model is implemented as a Recurrent State - Space Model (RSSM). Firstly, an encoder converts sensory inputs $x_t$ to stochastic representations $z_t$. Then, a sequence model with recurrent state $h_t$ forecasts the sequence of these representations given last actions $a_{t-1}$. The concatenation of $h_t$ and $z_t$ forms the model state from which we forecast rewards $r_t$ and episode continuation states $c_t \in \{0,1\}$ and reconstruct the inputs to ensure informative representations.

The actor and critic networks learn behaviors from abstract sequences imagined by the world model. The actor and critic operate on model states $s_t \doteq \{h_t, z_t\}$ and thus benefit from the rich representations learned by the world model.

## 3. TSC Environment

In this study, we utilized the single-agent corridor TSC model previously proposed by the author [9]. In this model, a traffic controller (Agent) learns an optimal policy to cooperative control the signal timing of multiple intersections on a corridor through interaction with the corridor simulation model (Environment). At an interaction step (simulation time is $t$), the traffic controller receives an observable state (State) from the simulation model and changes traffic signal schemes (Action) based on the state information and the policy. The traffic simulation model then proceeds to the next step (simulation time is $t + t_c$, where $t_c$ is a predefined control interval), and returns new traffic state information along with a reward.

The elements of RL for the TSC problem, including state space, action space, and reward function, are defined as follows.

### 3.1. State Space

The state space consists of two components: the current congestion state and the current signal phase scheme.

### 3.1.1. Current Congestion State

The congestion state is represented by the queue length for each link (section) at current time. The congestion state is a $L$ vector, where $L$ represents the number of links in a corridor.

$$[q_k^l] \qquad (1)$$

where $q_k^l$ is queue length on the link $l \in L$ at time $k$, $k$ is current time. The queue length is bounded by 0 and $q_{ub}$, which is predetermined.

The definition of queue length refers to the number of vehicles that are forced to stop at the straight lanes of a downstream intersection because of a red light. The queue length can be estimated using probe vehicle data and is closely related to the level of congestion [10].
.

### 3.1.2. Current Signal Phase Scheme

The current signal phase scheme is represented by signal phase split for each intersection of the corridor. This is a $M$ vector, where $M$ represents the number of signal intersections in a corridor:

$$[s_m] \qquad (2)$$

where $s_m$ represents the signal phase split at the intersection $m \in M$. The signal phase split is bounded by $s_{lb}$ and $s_{ub}$, and these values are predetermined.

In this model, each intersection is controlled by a four-phase signal (Figure 2). It is assumed that the signal cycle, left-turn phase time, yellow time, all-red time and offset time are predetermined and remain constant. The signal phase split is the only adjustable variable and defined as the sum of the north-south phases' time, which includes the north-south straight/right turn phase (Phase 1) and the left-turn phase (Phase 2). Since the signal cycle is constant, adjusting the signal phase split can control traffic flow at upstream and downstream intersections, thereby managing congestion levels. As illustrated in Figure 2, if congestion occurs on the link indicated by the arrow (from intersection $m$ to $m + 1$), the congestion can be alleviated by reducing the number of vehicles entering this link by increasing the signal phase split at the upstream intersection (intersection $m$), or/and by increasing the number of vehicles exiting this link by decreasing the signal phase split at the downstream intersection (intersection $m + 1$).

### 3.2. Action Space

For an intersection, the agent has three possible actions, and the action space is defined as $\{0,1,2\}$. Actions 0, 1, and 2 refer to adjusting the current signal phase split by $-\Delta s$, 0, and $\Delta s$, respectively, where $\Delta s$ is a predefined value.

For a corridor, the action space is a $M$ vector and can be represented as shown in Eq. (3), where $a_m$ denotes the action at the intersection $m \in M$.

$$[A_m] \qquad (3)$$

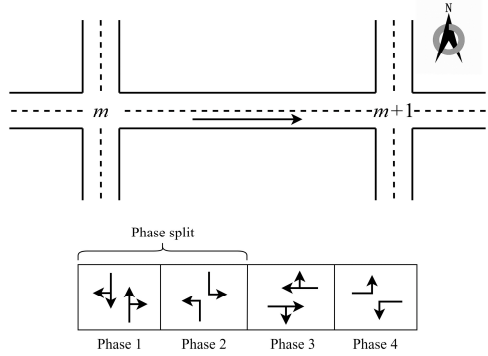

**Figure 2.** The signal phase split [9].

*3.3. Reward*

The reward is formulated as a function of the queue length for each link in a corridor.
The reward for a corridor is calculated as the sum of the rewards of its constituent links, with the reward for an individual link defined as shown in Eq. (4).

$$\begin{cases} q \leq q_{lc} : reward = 0 \\ q_{lc} \leq q \leq q_{hc}: reward = -(w_l \times q) \\ q \geq q_{hc}: reward = -(w_{cp} \times w_l \times q) \end{cases} \quad (4)$$

Where $q$ is the queue length, $q_{lc}$ and $q_{hc}$ are the thresholds for light and heavy congestion, respectively. $w_l$ represents the link importance weight and $w_{cp}$ is the penalty weight related to heavy congestion. It is included in cases of heavy congestion to prevent severe congestion from occurring. The values of $q_{lc}$, $q_{hc}$, $w_l$ and $w_{cp}$ are constants determined in advance.

## 4. Experiment Design

The aim of the experiments is to evaluate whether the recent successes of learned world models enable sample-efficient TSC strategy learning. Specifically, we intend to answer the following research questions:

- Does DreamerV3 enable TSC strategy learning?
- How do the two hyperparameters, model size and training ratio, affect the training performance?

*4.1. Simulation Model*

The experiments are carried out using SUMO, a sophisticated and adaptable traffic simulation software. SUMO features a dedicated interface known as libsumo, which allows users to control the simulation running, access real-time traffic data, and implement signal schemes in SUMO via Python scripts. The mesoscopic mode is utilized, which operates up to 100 times faster than the microscopic mode.

The geometry of this corridor is depicted in Figure 3. Each link has three lanes and extends to four lanes at the downstream intersection. Each approach consists of two straight lanes, one left-turn lane and one right-turn lane. Each intersection is controlled by four signal phases, with their sequence predetermined. The signal cycle, left-turn phase time, yellow time, and all-red time remain constant and are set to 100 seconds, 8 seconds, 2 seconds, and 2 seconds, respectively. The initial signal phase split is set to 50 seconds.

Traffic is generated using zone-to-zone demand generation and zone-to-zone flows are defined by an origin-destination (OD) matrix. Two patterns of OD matrix scenarios are tested. In the first pattern of the OD matrix (Scenario 1), the west-east bound flow is set to be dominant to simulate morning peak traffic in a corridor connecting suburban and central areas of a city. In the second pattern (Scenario 2), the east-west and west-east bound flows are set to be equal, simulating morning peak traffic in a corridor located in the central area of a city.

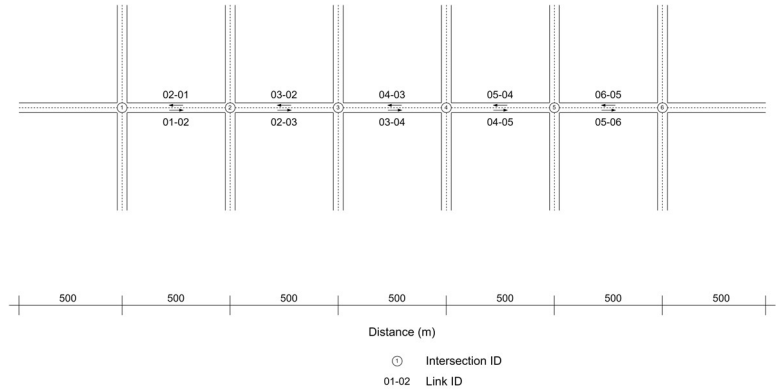

**Figure 3.** The Corridor Diagram.

*4.2. Parameter Setting*

Table 1 summarizes the parameters of RL-based TSC model.

When the signal phase split is set to its initial value (50 seconds), the saturated flow of the straight lanes within a signal cycle is estimated to be 50 vehicles. The upper bound of the queue length is set at this saturated flow level. This situation is considered an extreme congestion state because the queued vehicles occupy the entire green time, and all new incoming vehicles will encounter a red light and stop at the downstream intersection. The thresholds for light and heavy congestion ($q_{lc}$ and $q_{hc}$) are set at 20 % and 50 % of this saturated flow level.

In the first pattern of the OD matrix, where the major traffic flow is from west to east, the link importance weights ($w_l$) are set to 0 for the links in the opposite direction. This is done to consider only the links that are consistent with the major traffic flow in the reward calculation.

Table 1. RL-BASED TSC METHOD PARAMETERS

| Parameter | Value |
|---|---|
| $t_c$ | 100 s |
| $q_{ub}$ | 50 no. of vehicles |
| $q_{lc}$ and $q_{hc}$ | 10 and 25 no. of vehicles |
| $s_{lb}$ and $s_{ub}$ | 30 s and 70 s |
| $\Delta s$ | 2 s |
| $w_l$ | 0 for the east-west bound links in the first pattern of the OD matrix; 1 for otherwise |
| $w_{cp}$ | 10 |

The RL models are trained over multiple episodes, with each episode simulated for 16,200 seconds, including an 1,800-second warm-up period. During the warm-up, the signal scheme is unchanged, and the vehicles are incrementally introduced into the road network to establish a stable initial state. Each episode consists of 144 steps, calculated by dividing the remaining time after the warm-up period (16,200 - 1,800 seconds) by the control interval ($t_c$).

For DreamerV3, the Ray RLlib library is used to learn the TSC policy.

In this study, we employed a computing platform equipped with an Intel Core i9-13900K processor (8 Performance cores and 16 Efficient cores, totaling 32 threads) and an NVIDIA GeForce RTX 4090 24GB graphics card.

## 5. SIMULATION RESULTS AND DISCUSSIONS

*5.1. Hyperparameter Tuning for Dreamerv3*

For DreamerV3, two hyperparameters, namely the training ratio and model size, need to be adjusted. The training ratio is the ratio of replayed steps to environment steps, and thus, a higher training ratio results in substantially improved data efficiency.

For Scenario 1, Figure 4 illustrate the training process obtained by the DreamerV3 algorithm for various model sizes and training ratios. This study considered only the XS, S, M, and L size models; due to insufficient video memory, the XL size model was not included in this research. In these figures, the x-axis indicates the training hours, and the y-axis represents the episode reward. In order to better distinguish the performance of various hyperparameters at the end of training, the training curves after 4 hours were magnified and Figure 5 shows the average of episode reward after 5 hours of training.

Figure 5 shows that models of XS, S, M, and L sizes are all capable of identifying several training ratios suitable for Scenario 1. For example, the XS model is associated with training ratios of 64, 128 or 512, whereas the L model corresponds to a training ratio of 128, and similarly for the other sizes. However, model size S has the widest range of options for the selection of training ratio. As shown in Figure 4b, it can achieve good training results when the training ratio is between 64 and 512, and it exhibits excellent stability in the later stages of training. In contrast, model size L has a much narrower range of options for the training ratio. As shown in Figure 4d, it can only achieve satisfactory training results when the training ratio is at 128. Furthermore, as depicted in Figure 4, when a larger training ratio is chosen, the training curve exhibits significant fluctuations, and it does not fully stabilize by the end of training.

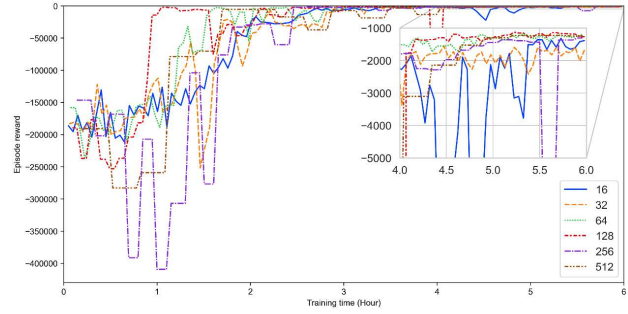

(a) Model size: XS.

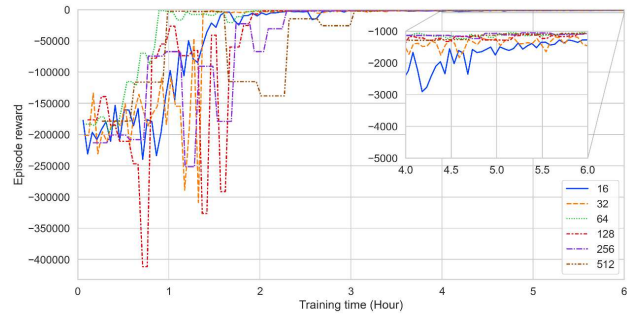

(b) Model size: S

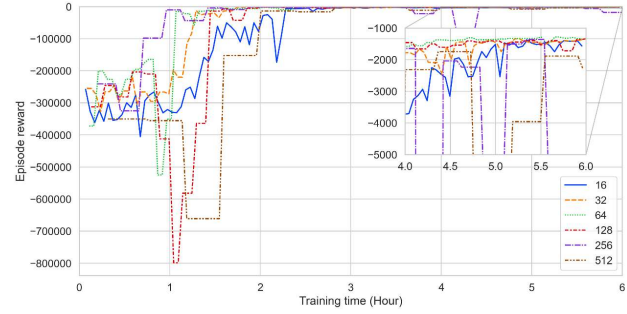

(c) Model size: M

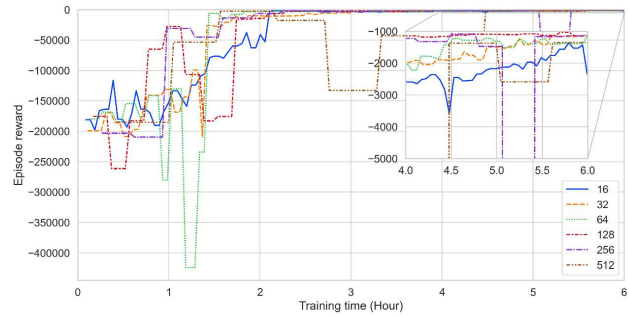

(d) Model size: L

**Figure 4.** The training process in Scenario 1.

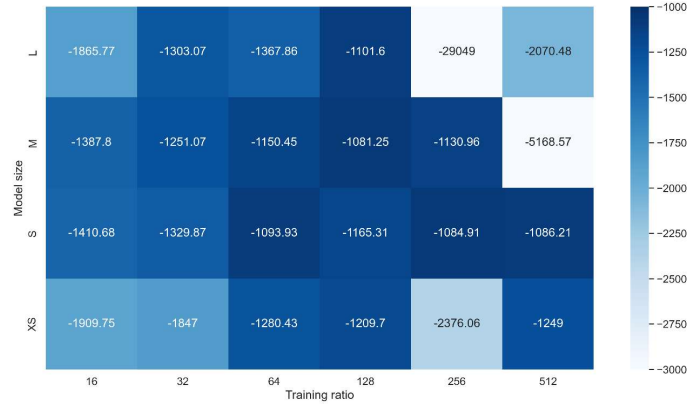

**Figure 5.** The training result in Scenario 1.

From the perspective of data efficiency, as the model size increases, it generally reaches a stable state earlier and maintains it until the later stages of training. For example, as shown in Figure 4, when the model size is XS, it takes about 3 hours of training to gradually stabilize, whereas for size L (except for training ratios of 256 and 512), it stabilizes in just a little over 2 hours. This indicates that larger model sizes have greater data efficiency than smaller models.

In summary, since the stability of the training process is more critical than data efficiency, for problems similar to those studied in this paper, if time for parameter tuning is limited, model size S should be prioritized.

From the perspective of the training ratio selection, it can be seen from the Figure 4 that a medium-sized training ratio yields better results, while the training curves with either too large or too small training ratios exhibit greater volatility. That is, if time for parameter tuning is limited, priority should be given to medium-sized training ratios between 64 to 512.

For Scenario 2, Figure 6 illustrates the training process obtained by the DreamerV3 algorithm for various model sizes and training ratios, and Figure 7 shows the average episode reward after 5 hours of training. From these figures, we can observe a situation similar to Scenario 1, where each model size is able to find several training ratios that are suitable for training in Scenario 2. However, from the perspective of the stability of the training process, model size S is the optimal choice and initially trying several medium training ratios can significantly reduce the time spent on hyperparameter tuning.

*5.2. Performance of TSC*

The performance of the Traffic Signal Control (TSC) model was evaluated by comparing it with the base case, where no adjustments were made to the signal timing schema,

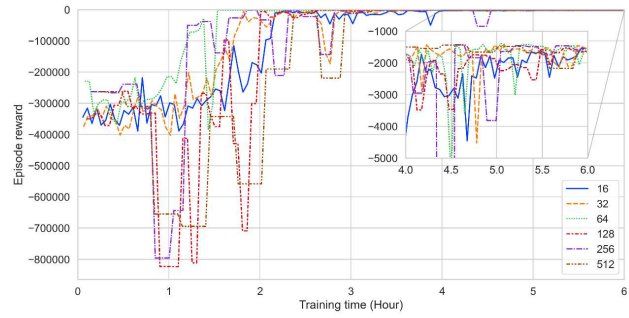
(a)   Model size: XS

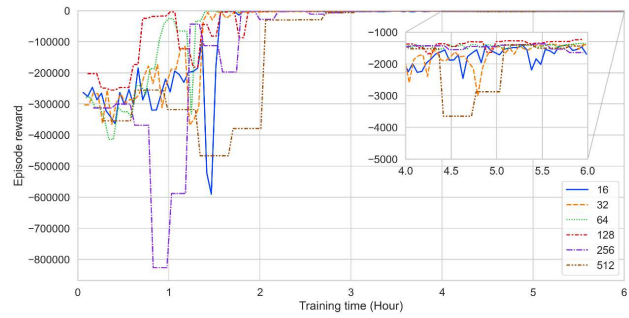
(b)   Model size: S

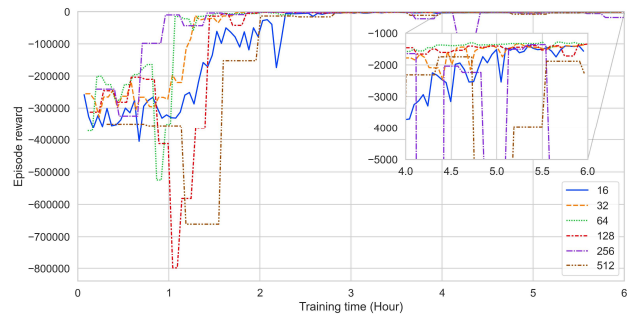
(c)   Model size: M

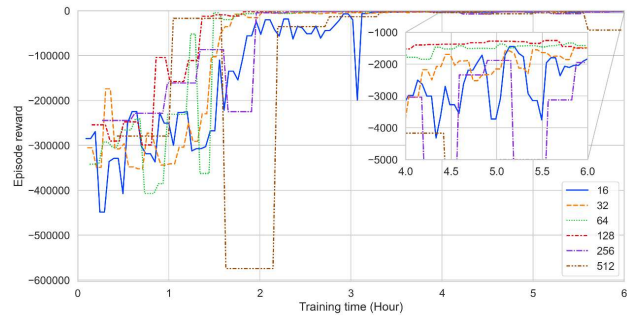
(d)   Model size: L

**Figure 6.** The training process in Scenario 2.

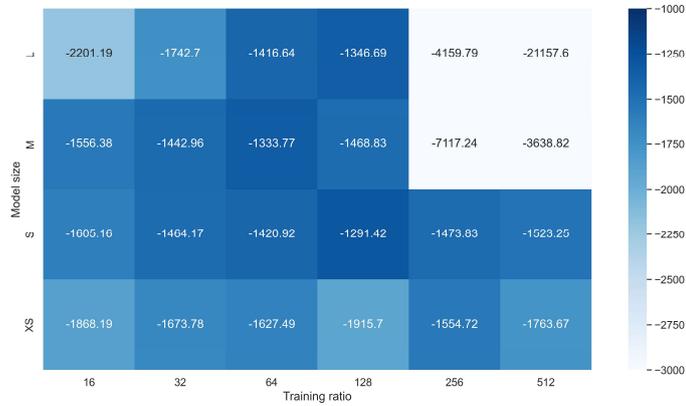

**Figure 7.** The training result in Scenario 2.

across two scenarios. The TSC model was trained using the DreamerV3 algorithm with a model size of S and a training ratio of 128 for both scenarios.

In scenario 1, Figure 8 depict the changes in queue length at each link throughout an episode. Only the queue lengths for links aligned with the main traffic flow direction are shown. In the base case, severe congestion is evident within the urban corridor. The queue lengths in many road sections exceed the set maximum of 50 and even reach over 100 in some instances. However, under the control of the proposed TSC model, the congestion levels are significantly reduced. This indicates that the proposed TSC method is highly effective in managing and preventing congestion in urban corridors, even when faced with significant traffic demand. It showcases the model's ability to optimize traffic flow and mitigate congestion-related issues.

In scenario 2, as shown in Figure 9, similar trends are observed. The base case again shows considerable congestion, while the TSC model trained with DreamerV3 successfully reduces the congestion levels. This consistency across both scenarios emphasizes the robustness and general applicability of the proposed TSC method.

The reduction in congestion levels not only leads to smoother traffic flow but also has implications for other aspects such as travel time reduction, fuel consumption savings, and environmental impact mitigation. By effectively managing the queue lengths at each link, the TSC model can prevent the formation of long queues and subsequent traffic jams. This, in turn, allows vehicles to move more freely and reduces the time spent idling, resulting in less fuel consumption and fewer emissions.

Overall, the performance of the TSC model trained with DreamerV3 demonstrates its potential as a valuable tool for traffic signal control. It offers a promising solution for improving traffic conditions in urban corridors and can be further explored and refined for more widespread application in real-world traffic management systems.

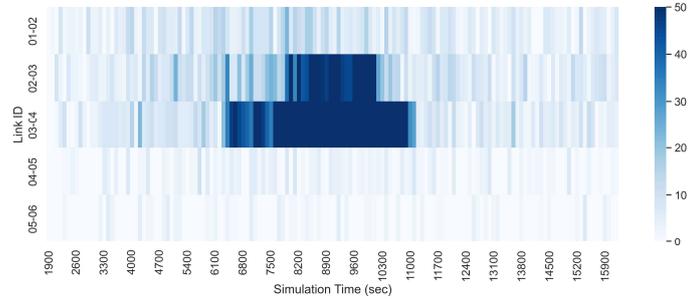

(a) Base case

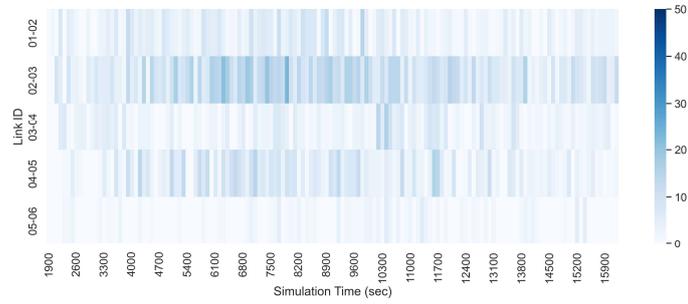

(b) Controlled case

**Figure 8.** Queue length comparisons in scenario 1.

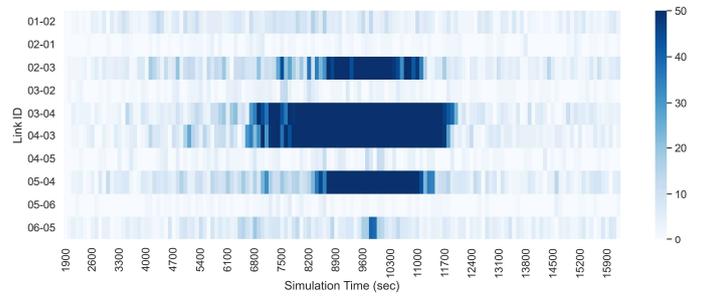

(a) Base case

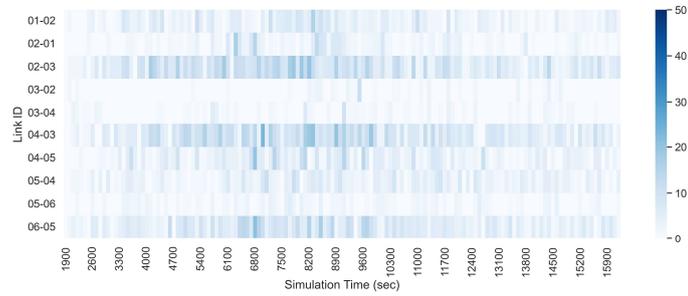

(b) Controlled case

**Figure 9.** Queue length comparisons in scenario 2.

faced with significant traffic demand. It showcases the model's ability to optimize traffic flow and mitigate congestion-related issues.

In scenario 2, as shown in Figure 9, similar trends are observed. The base case again shows considerable congestion, while the TSC model trained with DreamerV3 successfully reduces the congestion levels. This consistency across both scenarios emphasizes the robustness and general applicability of the proposed TSC method.

The reduction in congestion levels not only leads to smoother traffic flow but also has implications for other aspects such as travel time reduction, fuel consumption savings, and environmental impact mitigation. By effectively managing the queue lengths at each link, the TSC model can prevent the formation of long queues and subsequent traffic jams. This, in turn, allows vehicles to move more freely and reduces the time spent idling, resulting in less fuel consumption and fewer emissions.

Overall, the performance of the TSC model trained with DreamerV3 demonstrates its potential as a valuable tool for traffic signal control. It offers a promising solution for improving traffic conditions in urban corridors and can be further explored and refined for more widespread application in real-world traffic management systems.

## 6. Conclusion

In this research, we explored the application of DreamerV3 in Traffic Signal Control (TSC) learning. The DreamerV3 algorithm, with its unique properties, has shown potential in enabling sample-efficient policy learning, which is a significant advantage over traditional reinforcement learning algorithms that often require excessive interaction with the environment.

For DreamerV3, two hyperparameters, the model size and training ratio, play crucial roles and need to be carefully adjusted. Our experiments led to several important findings. Firstly, we found that selecting a model size of S and initially attempting several medium training ratios (such as those between 64 and 512) can substantially reduce the time spent on hyperparameter tuning. This is beneficial as it allows for more efficient model development. Moreover, we observed that this approach is generally applicable, as demonstrated by its ability to solve two distinct TSC task scenarios using the same set of hyperparameters (model size S and training ratio 128).

Regarding the claimed data-efficiency of the DreamerV3 algorithm, our analysis of the episode reward curve during the early stages of training provided some insights. Due to the significant fluctuations in this curve, we could only confirm that larger model sizes exhibit a certain level of data-efficiency. However, contrary to expectations, we found no evidence to suggest that increasing the training ratio accelerates convergence. In fact, an excessively high training ratio was found to lead to poorer training performance.

Looking ahead, there are several interesting research directions. One possibility is to expand the experiments to a broader range of road networks, including those with hundreds of intersections. This would provide a more comprehensive understanding of the DreamerV3 algorithm's performance in different traffic scenarios. Additionally, it would be valuable to compare the DreamerV3 algorithm with multi-agent reinforcement learning algorithms. Such a comparison could help to further clarify the strengths and weaknesses of DreamerV3 in the context of TSC and potentially lead to the development of more effective traffic control strategies.

## 7. Acknowledgments

This study was supported by Natural Science Foundation of Top Talent of SZTU (No. GDRC202322), Guangdong Province General Regular Projects of Social Sciences Planning (No. GD24CGL37), a grant from Department of Education of Guangdong Province (No. 2022KCXTD027) and the National Natural Science Foundation of China (No. 52472326).

**References**

[1] A. Haydari and Y. Yılmaz, "Deep Reinforcement Learning for Intelligent Transportation Systems: A Survey," IEEE Transactions on Intelligent Transportation Systems, vol. 23, no. 1, pp. 11–32, 2022, doi: 10.1109/TITS.2020.3008612.
[2] M. Noaeen et al., "Reinforcement learning in urban network traffic signal control: A systematic literature review," Expert Systems with Applications, vol. 199, p. 116830, Aug. 2022, doi: 10.1016/j.eswa.2022.116830.
[3] H. Wei, G. Zheng, V. Gayah, and Z. Li, "A Survey on Traffic Signal Control Methods." arXiv, Jan. 16, 2020. Accessed: Jul. 18, 2023. [Online]. Available: http://arxiv.org/abs/1904.08117
[4] H. Wei and G. Zheng, "Recent Advances in Reinforcement Learning for Traffic Signal Control," ACM SIGKDD Explorations Newsletter, vol. 22, pp. 12–18, 2021, [Online]. Available: https://api.semanticscholar.org/CorpusID:229540095
[5] D. Hafner, T. Lillicrap, J. Ba, and M. Norouzi, "Dream to Control: Learning Behaviors by Latent Imagination," ArXiv, 2020, [Online]. Available: https://arxiv.org/abs/1912.01603
[6] D. Hafner, T. Lillicrap, M. Norouzi, and J. Ba, "Mastering Atari with Discrete World Models." ArXiv, 2022. [Online]. Available: https://arxiv.org/abs/2010.02193
[7] D. Hafner, J. Pasukonis, J. Ba, and T. Lillicrap, "Mastering Diverse Domains through World Models," ArXiv, 2024, [Online]. Available: https://arxiv.org/abs/2301.04104
[8] D. R. Ha and J. Schmidhuber, "World Models," ArXiv, vol. abs/1803.10122, 2018, [Online]. Available: https://api.semanticscholar.org/CorpusID:4807711
[9] L. Liu, X. Zhuang, and Q. Li, "Adaptive Traffic Signal Control for Urban Corridor Based on Reinforcement Learning," in Computational and Experimental Simulations in Engineering, K. Zhou, Ed., Cham: Springer Nature Switzerland, 2025, pp. 25–35.
[10] Q. Li, H. Hu, and L. Miao, "Queue Length Estimation Using Probe Vehicle Data for a Congested Arterial Road," in CICTP 2015 - Efficient, Safe, and Green Multimodal Transportation - Proceedings of the 15th COTA International Conference of Transportation Professionals, Beijing, China, 2015, pp. 2295–2306. [Online]. Available: http://dx.doi.org/10.1061/9780784479292.213